\documentclass{article} 
\usepackage{ijcai20}

\usepackage{MnSymbol}%
\usepackage{wasysym}%
\usepackage{xcolor}
\usepackage{soul}
\usepackage{multirow}
\usepackage{url}
\usepackage[hidelinks]{hyperref}
\usepackage[utf8]{inputenc}
\usepackage[small]{caption}
\usepackage{graphicx}
\usepackage{subcaption}
\usepackage{amsmath}
\usepackage{amsthm}
\usepackage{booktabs}
\usepackage{algorithm}
\usepackage{algorithmic}
\usepackage{xcolor}
\usepackage{times}

\usepackage{hyperref}
\usepackage{url}
\usepackage{graphicx}
\usepackage{amsmath}
\usepackage{bbm}

\newtheorem{definition}{\textsc{Definition}}

\newcommand{\mb}{\mathbf}
\newcommand{\mc}{\mathcal}
\newcommand{\bs}{\boldsymbol}

\newcommand{\our}{\textsc{Graph-Bert}}
\newcommand{\bert}{\textsc{Bert}}
\newcommand{\transformer}{\textsc{Transformer}}

\newcommand{\gcn}{\textsc{GCN}}
\newcommand{\gat}{\textsc{GAT}}
\newcommand{\loopy}{\textsc{LoopyNet}}

\title{{\our}: Only Attention is Needed for Learning Graph Representations}

\author{Jiawei~Zhang$^\star$ \and Haopeng Zhang$^\star$ \and Congying Xia$^\dagger$ \and Li Sun$^\ddagger$\\
\affiliations
$^\star$IFM Lab, Florida State University, Tallahassee, FL, USA\\
$^\dagger$University of Illinois at Chicago, IL, USA\\
$^\ddagger$Beijing University of Posts and Telecommunications, Beijing, China\\
 \emails
 \{ jiawei, haopeng \}@ifmlab.org, cxia8@uic.edu, l.sun@bupt.edu.cn}

\begin{document}

\maketitle

\begin{abstract}

The dominant graph neural networks (GNNs) over-rely on the graph links, several serious performance problems with which have been witnessed already, e.g., \textit{suspended animation problem} and \textit{over-smoothing problem}. What's more, the inherently inter-connected nature precludes parallelization within the graph, which becomes critical for large-sized graph, as memory constraints limit batching across the nodes. In this paper, we will introduce a new graph neural network, namely {\our} (Graph based {\bert}), solely based on the attention mechanism without any graph convolution or aggregation operators. Instead of feeding {\our} with the complete large input graph, we propose to train {\our} with sampled linkless subgraphs within their local contexts. {\our} can be learned effectively in a standalone mode. Meanwhile, a pre-trained {\our} can also be transferred to other application tasks directly or with necessary fine-tuning if any supervised label information or certain application oriented objective is available. We have tested the effectiveness of {\our} on several graph benchmark datasets. Based the pre-trained {\our} with the \textit{node attribute reconstruction} and \textit{structure recovery} tasks, we further fine-tune {\our} on \textit{node classification} and \textit{graph clustering} tasks specifically. The experimental results have demonstrated that {\our} can out-perform the existing GNNs in both the learning effectiveness and efficiency. 

\end{abstract}
\section{Introduction}\label{sec:introduction}


Graph provides a unified representation for many inter-connected data in the real-world, which can model both the diverse attribute information of the node entities and the extensive connections among these nodes. For instance, the human brain imaging data, online social media and bio-medical molecules can all be represented as graphs, i.e., the brain graph \cite{Meng_Isomorphic_19}, social graph \cite{Ugander_Anatomy_11} and molecular graph \cite{Jin_Junction_18}, respectively. Traditional machine learning models can hardly be applied to the graph data directly, which usually take the feature vectors as the inputs. Viewed in such a perspective, learning the representations of the graph structured data is an important research task.

In recent years, great efforts have been devoted to designing new graph neural networks (GNNs) for effective graph representation learning. Besides the network embedding models, e.g., node2vec \cite{DBLP:journals/corr/GroverL16} and deepwalk \cite{Perozzi:2014:DOL:2623330.2623732}, the recent graph neural networks, e.g., {\gcn} \cite{Kipf_Semi_CORR_16}, {\gat} \cite{Velickovic_Graph_ICLR_18} and {\loopy} \cite{loopynet}, are also becoming much more important, which can further refine the learned representations for specific application tasks. Meanwhile, most of these existing graph representation learning models are still based on the graph structures, i.e., the links among the nodes. Via necessary neighborhood information aggregation or convolutional operators along the links, nodes' representations learned by such approaches can preserve the graph structure information. 

However, several serious learning performance problem, e.g., suspended animation problem \cite{Zhang2019GResNetGR} and over-smoothing problem \cite{Li_Deeper_CORR_18}, with the existing GNN models have also been witnessed in recent years. According to \cite{Zhang2019GResNetGR}, for the GNNs based on the approximated graph convolutional operators \cite{Hammond_2011}, as the model architecture goes deeper and reaches certain limit, the model will not respond to the training data and suffers from the suspended animation problem. Meanwhile, the node representations obtained by such deep models tend to be over-smoothed and also become indistinguishable \cite{Li_Deeper_CORR_18}. Both of these two problems greatly hinder the applications of GNNs for deep graph representation learning tasks. What's more, the inherently inter-connected nature precludes parallelization within the graph, which becomes critical for large-sized graph input, as memory constraints limit batching across the nodes.

To address the above problems, in this paper, we will propose a new graph neural network model, namely {\our} (Graph based {\bert}). Inspired by \cite{ZCG18}, model {\our} will be trained with sampled nodes together with their context (which are called linkless subgraphs in this paper) from the input large-sized graph data. Distinct from the existing GNN models, in the representation learning process, {\our} utilizes no links in such sampled batches, which will be purely based on the attention mechanisms instead \cite{VSPUJGKP17,DCLT18}. Therefore, {\our} can get rid of the aforementioned learning effectiveness and efficiency problems with existing GNN models promisingly.

What's more, compared with computer vision \cite{DBLP:journals/corr/abs-1811-08883} and natural language processing \cite{DCLT18}, graph neural network pre-training and fine-tuning are still not common practice by this context so far. The main obstacles that prevent such operations can be due to the diverse input graph structures and the extensive connections among the nodes. Also the different learning task objectives also prevents the transfer of GNNs across different tasks. Since {\our} doesn't really rely on the graph links at all, in this paper, we will investigate the transfer of pre-trained {\our} on new learning tasks and other sequential models (with necessary fine-tuning), which will also help construct the functional pipeline of models in graph learning.

We summarize our contributions of this paper as follows:
\begin{itemize}

\item \textbf{New GNN Model}: In this paper, we introduce a new GNN model {\our} for graph data representation learning. {\our} doesn't rely on the graph links for representation learning and can effectively address the suspended animation problems aforementioned. Also {\our} is trainable with sampled linkless subgraphs (i.e., target node with context), which is more efficient than existing GNNs constructed for the complete input graph. To be more precise, the training cost of {\our} is only decided by (1) training instance number, and (2) sampled subgraph size, which is uncorrelated with the input graph size at all.

\item \textbf{Unsupervised Pre-Training}: Given the input unlabeled graph, we will pre-train {\our} based on to two common tasks in graph studies, i.e., node attribute reconstruction and graph structure recovery. Node attribute recovery ensures the learned node representations can capture the input attribute information; whereas graph structure recovery can further ensure {\our} learned with linkless subgraphs can still maintain both the graph local and global structure properties. 

\item \textbf{Fine-Tuning and Transfer}: Depending on the specific application task objectives, the {\our} model can be further fine-tuned to adapt the learned representations to specific application requirements, e.g., node classification and graph clustering. Meanwhile, the pre-trained {\our} can also be transferred and applied to other sequential models, which allows the construction of functional pipelines for graph learning.

\end{itemize}

The remaining parts of this paper are organized as follows. We will introduce the related work in Section~\ref{sec:related_work}. Detailed information about the {\our} model will be introduced in Section~\ref{sec:method}, whereas the pre-training and fine-tuning of {\our} will be introduced in Section~\ref{sec:analysis} in detail. The effectiveness of {\our} will be tested in Section~\ref{sec:experiment}. Finally, we will conclude this paper in Section~\ref{sec:conclusion}.

\section{Related Work}\label{sec:related_work}

To make this paper self-contained, we will introduce some related topics here on {GNNs}, {\transformer} and {\bert}.

\vspace{5pt}

\noindent \textbf{Graph Neural Network}: Representative examples of GNNs proposed by present include {\gcn} \cite{Kipf_Semi_CORR_16}, GraphSAGE \cite{Hamilton_Inductive_17} and {\loopy} \cite{loopynet}, based on which various extended models \cite{Velickovic_Graph_ICLR_18,sun2019adagcn,DBLP:journals/corr/abs-1810-05997} have been introduced as well. As mentioned above, {\gcn} and its variant models are all based on the approximated graph convolutional operator \cite{Hammond_2011}, which may lead to the suspended animation problem \cite{Zhang2019GResNetGR} and over-smoothing problem \cite{Li_Deeper_CORR_18} for deep model architectures. Theoretic analyses of the reasons are provided in \cite{Li_Deeper_CORR_18,Zhang2019GResNetGR,Merve_An_19}. To handle such problems, \cite{Zhang2019GResNetGR} generalizes the graph raw residual terms in \cite{loopynet} and proposes a method based on graph residual learning; \cite{Li_Deeper_CORR_18} proposes to adopt residual/dense connections and dilated convolutions into the GCN architecture. Several other works \cite{sun2019adagcn,Huang_Inductive_19} seek to involve the recurrent network for deep graph representation learning instead.

\vspace{5pt}

\noindent \textbf{{\bert} and {\transformer}}: In NLP, the dominant sequence transduction models are based on complex recurrent \cite{Hochreiter_Long_Neural_97,DBLP:journals/corr/ChungGCB14} or convolutional neural networks \cite{kim-2014-convolutional}. However, the inherently sequential nature precludes parallelization within training examples. Therefore, in \cite{VSPUJGKP17}, the authors propose a new network architecture, the {\transformer}, based solely on attention mechanisms, dispensing with recurrence and convolutions entirely. With {\transformer}, \cite{DCLT18} further introduces {\bert} for deep language understanding, which obtains new state-of-the-art results on eleven natural language processing tasks. In recent years, {\transformer} and {\bert} based learning approaches have been used extensively in various learning tasks \cite{DBLP:journals/corr/abs-1901-02860,lan2019albert,DBLP:journals/corr/abs-1906-00346}. 

\vspace{5pt}

Readers may also refer to page\footnote{https://paperswithcode.com/area/graphs} and page\footnote{https://paperswithcode.com/area/natural-language-processing} for more information on the state-of-the-art work on these topics.
\begin{figure*}
    \centering
    \begin{minipage}{.95\textwidth}
    	\includegraphics[width=\linewidth]{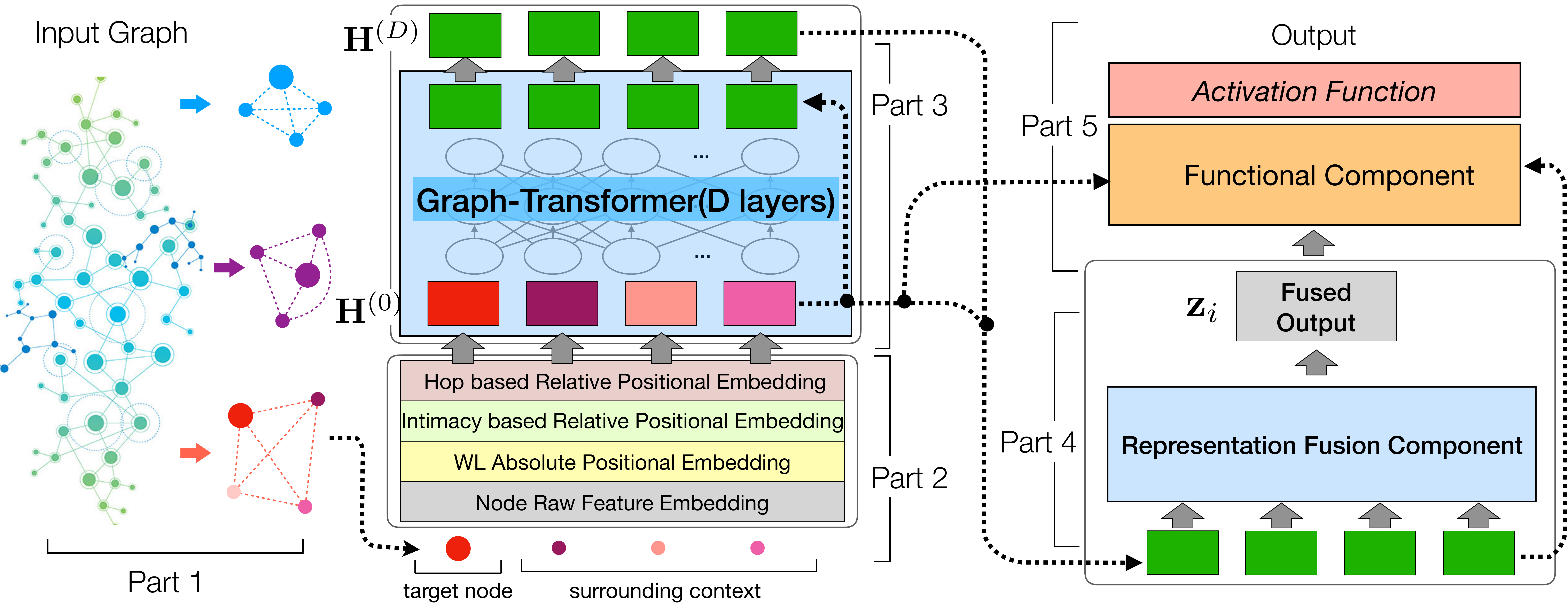}
     \end{minipage}%
        \caption{Architecture of the {\our} Model. (Part 1: linkless subgraph batching; Part 2: node input vector embeddings; Part 3: graph transformer based encoder; Part 4: representation fusion; Part 5: functional component. Depending on the target application task, the function component will generate different output. In the sampled subgraphs, it covers both the target node and the surrounding context nodes.)}
    	\label{fig:framework}
\end{figure*}

\section{Method}\label{sec:method}

In this section, we will introduce the detailed information about the {\our} model. As illustrated in Figure~\ref{fig:framework}, {\our} involves several parts: (1) linkless subgraph batching, (2) node input embedding, (3) graph-transformer based encoder, (4) representation fusion, and (5) the functional component. The results learned by the the graph-transformer model will be fused as the representation for the target nodes. In this section, we will introduce these key parts in great detail, whereas the pre-training and fine-tuning of {\our} will be introduced in the following section.

\subsection{Notations}

In the sequel of this paper, we will use the lower case letters (e.g., $x$) to represent scalars, lower case bold letters (e.g., $\mb{x}$) to denote column vectors, bold-face upper case letters (e.g., $\mb{X}$) to denote matrices, and upper case calligraphic letters (e.g., $\mathcal{X}$) to denote sets or high-order tensors. Given a matrix $\mb{X}$, we denote $\mb{X}(i,:)$ and $\mb{X}(:,j)$ as its $i_{th}$ row and $j_{th}$ column, respectively. The ($i_{th}$, $j_{th}$) entry of matrix $\mb{X}$ can be denoted as either $\mb{X}(i,j)$. We use $\mb{X}^\top$ and $\mb{x}^\top$ to represent the transpose of matrix $\mb{X}$ and vector $\mb{x}$. For vector $\mb{x}$, we represent its $L_p$-norm as $\left\| \mb{x} \right\|_p = (\sum_i |\mb{x}(i)|^p)^{\frac{1}{p}}$. The Frobenius-norm of matrix $\mb{X}$ is represented as $\left\| \mb{X} \right\|_F = (\sum_{i,j} |\mb{X}(i,j)|^2)^{\frac{1}{2}}$. The element-wise product of vectors $\mb{x}$ and $\mb{y}$ of the same dimension is represented as $\mb{x} \otimes \mb{y}$, whose concatenation is represented as $\mb{x} \sqcup \mb{y}$.


\subsection{Linkless Subgraph Batching}\label{subsec:sampling}

Prior to talking about the subgraph batching method, we would like to present the problem settings first. Formally, we can represent the input graph data as $G = (\mc{V}, \mc{E}, w, x, y)$, where $\mc{V}$ and $\mc{E}$ denote the sets of nodes and links in graph $G$, respectively. Mapping $w: \mc{E} \to \mathbbm{R}$ projects links to their weight; whereas mappings $x: \mc{V} \to \mc{X}$ and $y: \mc{V} \to \mc{Y}$ can project the nodes to their raw features and labels. The graph size can be represented by the number of involved nodes, i.e., $|\mc{V}|$. The above term defines a general graph concept. If the studied $G$ is unweighted, we will have $w(v_i, v_j) = 1, \forall (v_i, v_j) \in \mc{E}$; whereas $\forall (v_i, v_j) \in \mc{V} \times \mc{V} \setminus \mc{E}$, we have $w(v_i, v_j) = 0$. Notations $\mc{X}$ and $\mc{Y}$ denote feature space and label space, respectively. In this paper, we can simply represent $\mc{X} = \mathbbm{R}^{d_x}$ and $\mc{Y} = \mathbbm{R}^{d_y}$. For node $v_i$, we can also simplify its raw feature and label vector representations as $\mb{x}_i = x(v_i) \in \mathbbm{R}^{d_x}$ and $\mb{y}_i = y(v_i) \in \mathbbm{R}^{d_y}$. The {\our} model pre-training doesn't require any label supervision information actually, but partial of the labels will be used for the fine-tuning application task on node classification to be introduced later.

Instead of working on the complete graph $G$, {\our} will be trained with linkless subgraph batches sampled from the input graph instead. It will effectively enable the learning of {\our} to parallelize (even though we will not study parallel computing of {\our} in this paper) on extremely large-sized graphs that the existing graph neural networks cannot handle. Different approaches can be adopted here to sample the subgraphs from the input graph as studied in\cite{ZCG18}. However, to control the randomness involved in the sampling process, in this paper, we introduce the \textit{top-k intimacy} sampling approach instead. Such a sampling algorithm works based on the graph intimacy matrix $\mb{S} \in \mathbbm{R}^{|\mc{V}| \times |\mc{V}|}$, where entry $\mb{S}(i, j)$ measures the intimacy score between nodes $v_i$ and $v_j$.

There exist different metrics to measure the intimacy scores among the nodes within the graph, e.g., Jaccard's coefficienty \cite{jaccard1901etude}, Adamic/Adar \cite{adamic2003friends}, Katz \cite{Katz1953}. In this paper, we define matrix $\mb{S}$ based on the pagerank algorithm, which can be denoted as
\begin{equation}
\mb{S} = \alpha \cdot \left(  \mb{I} - (1-\alpha) \cdot \bar{\mb{A}}  \right)^{-1},
\end{equation}
where factor $\alpha \in [0, 1]$ (which is usually set as $0.15$). Term $\bar{\mb{A}} = \mb{A} \mb{D}^{-1}$ denotes the colum-normalized adjacency matrix. In its representation, $\mb{A}$ is the adjacency matrix of the input graph, and $\mb{D}$ is its corresponding diagonal matrix with $\mb{D}(i,i) = \sum_j \mb{A}(i,j)$ on its diagonal. 

Formally, for any target node $v_i \in \mc{V}$ in the input graph, based on the intimacy matrix $\mb{S}$, we can define its learning context as follows:
\begin{definition}
(Node Context): Given an input graph $G$ and its intimacy matrix $\mb{S}$, for node $v_i$ in the graph, we define its learning context as set $\Gamma(v_i) = \{v_j | {v_j \in \mc{V} \setminus \{v_i\}} \land {\mb{S}(i,j) \ge \theta_i }\}$. Here, the term $\theta_i$ defines the minimum intimacy score threshold for nodes to involve in $v_i$'s context. 
\end{definition}

We may need to add a remark: for all the nodes in $v_i$' learning context $\Gamma(v_i)$, they can cover both local neighbors of $v_i$ as well as the nodes which are far away. In this paper, we define the threshold $\theta_i$ as the $k_{th}$ entry of $\mbox{sorted}(\mb{S}(i,:))$ (with $v_i$ being excluded), i.e., $\Gamma(v_i)$ covers the \textit{top-k} intimate nodes of $v_i$ in graph $G$. Based on the node context concept, we can also represent the set of sampled graph batches for all the nodes as set $\mc{G} = \{g_1, g_2, \cdots, g_{|\mc{V}|}\}$, and $g_i$ denotes the subgraph sampled for $v_i$ (as the target node). Formally, $g_i$ can be represented as $g_i = (\mc{V}_i, \emptyset)$, where the node set $\mc{V}_i = \{v_i\} \cup \Gamma(v_i)$ covers both $v_i$ and its context nodes and the link set is null. For large-sized input graphs, set $\mc{G}$ can further be decomposed into several mini-batches, i.e., $\mc{B} \subseteq \mc{G}$, which will be fed to train the {\our} model.


\subsection{Node Input Vector Embeddings}

Different from image and text data, where the pixels and words/chars have their inherent orders, nodes in graphs are orderless. The {\our} model to be learned in this paper doesn't require any node orders of the input sampled subgraph actually. Meanwhile, to simplify the presentations, we still propose to serialize the input subgraph nodes into certain ordered list instead. Formally, for all the nodes $\mc{V}_i$ in the sampled linkless subgraph $g_i \in \mc{B}$, we can denote them as a node list $[v_i, v_{i,1}, \cdots, v_{i,k}]$, where $v_{i,j}$ will be placed ahead of $v_{i,m}$ if $\mb{S}(i,j) > \mb{S}(i,m), \forall v_{i,j}, v_{i,m} \in \mc{V}_i$. For the remaining of this subsection, we will follow the identical node orders as indicated above by default to define their input vector embeddings.

The input vector embeddings to be fed to the graph-transformer model actually cover four parts: (1) raw feature vector embedding, (2) Weisfeiler-Lehman absolute role embedding, (3) intimacy based relative positional embedding, and (4) hop based relative distance embedding, respectively. 


\subsubsection{Raw Feature Vector Embedding}

Formally, for each node $v_j \in \mc{V}_i$ in the subgraph $g_i$, we can embed its raw feature vector into a shared feature space (of the same dimension $d_h$) with its raw feature vector $\mb{x}_j$, which can be denoted as
\begin{equation}
{\mb{e}}_{j}^{(x)} = \mbox{Embed} \left( \mb{x}_j\right) \in \mathbbm{R}^{d_h \times 1}.
\end{equation}
Depending on the input raw features properties, different models can be used to define the $\mbox{Embed}(\cdot)$ function. For instance, CNN can be used if $\mb{x}_j$ denotes images; LSTM/BERT can be applied if $\mb{x}_j$ denotes texts; and simple fully connected layers can also be used for simple attribute inputs.

\subsubsection{Weisfeiler-Lehman Absolute Role Embedding}

The Weisfeiler-Lehman (WL) algorithm \cite{DBLP:journals/corr/NiepertAK16} can label the nodes according to their structural roles in the graph data, where the nodes with the identical roles will be labeled with the same code (e.g., integer strings or node colors). Formally, for node $v_j \in \mc{V}_i$ in the sampled subgraph, we can denote its WL code as $\mbox{WL}(v_j) \in \mathbbm{N}$, which can be pre-computed based on the complete graph and is invariant for different sampled subgraphs. In this paper, we adopt the embedding approach proposed in \cite{VSPUJGKP17} and define the nodes WL absolute role embedding vector as
\begin{equation}
\begin{aligned}
{\mb{e}}_{j}^{(r)}& = \mbox{Position-Embed} \left( \mbox{WL}(v_j) \right)\\
&=  \left[sin\left (\frac{\mbox{WL}(v_j)}{10000^{\frac{2 l}{d_{h}}}} \right), cos\left(\frac{\mbox{WL}(v_j)}{10000^{\frac{2 l + 1}{d_{h}}}} \right) \right]_{l=0}^{\left \lfloor \frac{d_h}{2} \right \rfloor},
\end{aligned}
\end{equation}
where ${\mb{e}}_{j}^{(r)} \in \mathbbm{R}^{d_h \times 1}$. The index $l$ iterates throughout all the entries in the above vector to compute the entry values with $sin(\cdot)$ and $cos(\cdot)$ functions for the node based on its WL code.

\subsubsection{Intimacy based Relative Positional Embedding}

The WL based role embeddings can capture the global node role information in the representations. Here, we will introduce a relative positional embedding to extract the local information in the subgraph based on the placement orders of the serialized node list introduced at the beginning of this subsection. Formally, based on that serialized node list, we can denote the position of $v_j \in \mc{V}_i$ as $P(v_j)$. We know that $P(v_i) = 0$ by default and nodes closer to $v_i$ will have a small positional index. Furthermore, $P(\cdot)$ is a variant position index metric. For the identical node $v_j$, its positional index $P(v_j)$ will be different for different sampled subgraphs.

Formally, for node $v_j$, we can also extract its intimacy based relative positional embedding with the $\mbox{Position-Embed}(\cdot)$ function defined above as follows:
\begin{equation}
{\mb{e}}_{j}^{(p)} = \mbox{Position-Embed} \left( \mbox{P}(v_j) \right) \in \mathbbm{R}^{d_h \times 1},
\end{equation}
which is quite close to the positional embedding in \cite{VSPUJGKP17} for the relative positions in the word sequence.

\vspace{-3pt}
\subsubsection{Hop based Relative Distance Embedding}

The hop based relative distance embedding can be treated as a balance between the absolute role embedding (for global information) and intimacy based relative positional embedding (for local information). Formally, for node $v_j \in \mc{V}_i$ in the subgraph $g_i$, we can denote its relative distance in hops to $v_i$ in the original input graph as $H(v_j; v_i)$, which can be used to define its embedding vector as
\begin{equation}
{\mb{e}}_{j}^{(d)} = \mbox{Position-Embed} \left( \mbox{H}(v_j; v_i) \right) \in \mathbbm{R}^{d_h \times 1}.
\end{equation}
It it easy to observe that vector ${\mb{e}}_{j}^{(d)}$ will also be variant for the identical node $v_j$ in different subgraphs.


\subsection{Graph Transformer based Encoder}

Based on the computed embedding vectors defined above, we will be able to aggregate them together to define the initial input vectors for nodes, e.g., $v_j$, in the subgraph $g_i$ as follows: 
\begin{equation}
\mb{h}_j^{(0)} = \mbox{Aggregate} ({\mb{e}}_{j}^{(x)}, {\mb{e}}_{j}^{(r)}, {\mb{e}}_{j}^{(p)}, {\mb{e}}_{j}^{(d)}).
\end{equation}
In this paper, we simply define the aggregation function as the vector summation. Furthermore, the initial input vectors for all the nodes in $g_i$ can organized into a matrix $\mb{H}^{(0)} = [\mb{h}_i^{(0)}, \mb{h}_{i,1}^{(0)}, \cdots, \mb{h}_{i,k}^{(0)}]^\top \in \mathbbm{R}^{(k+1) \times d_h}$. The graph-transformer based encoder to be introduced below will update the nodes' representations iteratively with multiple layers ($D$ layers), and the output by the $l_{th}$ layer can be denoted as\vspace{-4pt}
\begin{equation}
\begin{aligned}
\hspace{-5pt} \mb{H}^{(l)} &= \mbox{G-Transformer} \left( \mb{H}^{(l-1)}\right)\\
&= \mbox{softmax} \left(\frac{\mb{Q} \mb{K}^\top}{\sqrt{d_h}} \right) \mb{V} + \mbox{G-Res} \left( \mb{H}^{(l-1)}, \mb{X}_i\right),
\end{aligned}\vspace{-2pt}
\end{equation}
where
\begin{equation}
\begin{cases}
\mb{Q} & = \mb{H}^{(l-1)} \mb{W}_Q^{(l)},\\
\mb{K} & = \mb{H}^{(l-1)} \mb{W}_K^{(l)},\\
\mb{V} & = \mb{H}^{(l-1)} \mb{W}_V^{(l)}.\\
\end{cases}\vspace{-2pt}
\end{equation}
In the above equations, $\mb{W}_Q^{(l)}, \mb{W}_K^{(l)}, \mb{W}_K^{(l)} \in \mathbbm{R}^{d_h \times d_h}$ denote the involved variables. To simplify the presentations in the paper, we assume nodes' hidden vectors in different layers have the same length. Notation $\mbox{G-Res} \left( \mb{H}^{(l-1)}, \mb{X}_i\right)$ represents the graph residual term introduced in \cite{Zhang2019GResNetGR}, and $\mb{X}_i \in \mathbbm{R}^{(k+1) \times d_x}$ is the raw features of all nodes in the subgraph $g_i$. Also different from conventional residual learning, we will add the residual terms computed for the target node $v_i$ to the hidden state vectors of all the nodes in the subgraph at each layer. Based on the graph-transformer function defined above, we can represent the representation learning process of {\our} as follows:
\begin{equation}
\begin{cases}
\vspace{5pt}
\mb{H}^{(0)} & \hspace{-10pt} = [\mb{h}_i^{(0)}, \mb{h}_{i,1}^{(0)}, \cdots, \mb{h}_{i,k}^{(0)}]^\top ,\\
\vspace{5pt}
\mb{H}^{(l)} &\hspace{-10pt}= \mbox{G-Transformer} \left( \mb{H}^{(l-1)}\right), \forall l \in \{1, 2, \cdots, D\},\\
\mb{z}_i &\hspace{-10pt}= \mbox{Fusion} \left( \mb{H}^{(D)} \right). 
\end{cases}
\end{equation}
Different from the application of conventional transformer model on NLP problems, which aims at learning the representations of all the input tokens. In this paper, we aim to apply the graph-transformer to get the representations of the target node only. In the above equation, function $\mbox{Fusion} \left(\cdot\right)$ will average the representations of all the nodes in input list, which defines the final state of the target $v_i$, i.e., $\mb{z}_i \in \mathbbm{R}^{d_h \times 1}$. Both vector $\mb{z}_i$ and matrix $\mb{H}^{(D)}$ will be outputted to the following functional component attached to {\our}. Depending on the application tasks, the functional component and learning objective (i.e., the loss function) will be different. We will show more detailed information in the following section on {\our} pre-training and fine-tuning.

\section{{\our} Learning}\label{sec:analysis}

We propose to pre-train {\our} with two tasks: (1) node attribute reconstruction, and (2) graph structure recovery. Meanwhile, depending on the objective application tasks, e.g., (1) node classification and (2) graph clustering as studied in this paper, {\our} can be further fine-tuned to adapt both the model and the learned node representations accordingly to the new tasks.

\subsection{Pre-training}

The node raw attribute reconstruction task focuses on capturing the node attribute information in the learned representations, whereas the graph structure recovery task focuses more on the graph connection information instead.

\subsubsection{Task \#1: Node Raw Attribute Reconstruction}

Formally, for the target node $v_i$ in the sampled subgraph $g_i$, we have its learned representation by {\our} to be $\mb{z}_i$. Via the fully connected layer (together with the activation function layer if necessary), we can denote the reconstructed raw attributes for node $v_i$ based on $\mb{z}_i$ as $\hat{\mb{x}}_i = \mbox{FC}(\mb{z}_i)$. To ensure the learned representations can capture the node raw attribute information, compared against the node raw features, e.g., $\mb{x}_i$ for $v_i$, we can define the node raw attribute reconstruction based loss term as follows:
\begin{equation}
\ell_{1} = \frac{1}{|\mc{V}|} \sum_{v_i \in \mc{V}} \left\| \mb{x}_i - \hat{\mb{x}}_i  \right\|_2.
\end{equation}

\subsubsection{Task \#2: Graph Structure Recovery}

Furthermore, to ensure such representation vectors can also capture the graph structure information, the graph structure recovery task is also used as a pre-training task. Formally, for any two nodes $v_i$ and $v_j$, based on their learned representations, we can denote the inferred connection score between them by computing their cosine similarity, i.e., $\hat{s}_{i,j} = \frac{\mb{z}_i^\top \mb{z}_j}{ \left\|  \mb{z}_i \right\| \left\| \mb{z}_j \right\|}$. Compared against the ground truth graph intimacy matrix defined in Section~\ref{subsec:sampling}, i.e., ${\mb{S}}$, we can denote the introduced loss term as follows:
\begin{equation}
\ell_2 =\frac{1}{|\mc{V}|^2} \left\| \mb{S} -\hat{\mb{S}} \right\|_F^2,
\end{equation}
where $\hat{\mb{S}} \in \mathbbm{R}^{|\mc{V}| \times |\mc{V}|}$ with entry $\hat{\mb{S}}(i,j) = \hat{s}_{i,j}$.

\subsection{Model Transfer and Fine-tuning}

In applying the learned {\our} into new learning tasks, the learned graph representations can be either fed into the new tasks directly or with necessary adjustment, i.e., fine-tuning. In this part, we can take the \textit{node classification} and \textit{graph clustering} tasks as the examples, where \textit{graph clustering} can use the learned representations directly but fine-tuning will be necessary for the \textit{node classification} task.

\subsubsection{Task \# 1: Node Classification}\label{subsec:node_classification}

Based on the nodes learned representations, e.g., $\mb{z}_i$ for $v_i$, we can denote the inferred label for the node via the functional component as $\hat{\mb{y}}_i = \mbox{softmax}( \mbox{FC} (\mb{z}_i))$. Compared with the nodes' true labels, we will be able to define the introduced node classification loss term on training batch $\mc{T}$ as
\begin{equation}
\ell_{nc} = \sum_{v_i \in \mc{T}} \sum_{m = 1}^{d_y} - \mb{y}_i(m) \log \hat{\mb{y}}_i (m).
\end{equation}
By re-training these stacked fully connected layers together with {\our} (loaded from pre-training), we will be able to infer node class labels.

\subsubsection{Task \# 2: Graph Clustering}\label{subsec:graph_clustering}

Meanwhile, for the graph clustering task, the main objective is to partition nodes in the graph into several different clusters, e.g., $\mc{C} = \{\mc{C}_1, \mc{C}_2, \cdots, \mc{C}_l\}$ ($l$ is a hyper-parameter pre-specified in advance). For each objective cluster, e.g., $\mc{C}_j \in \mc{C}$, we can denote its center as a variable vector $\bs{\mu}_j = \sum_{v_i \in\mc{C}_j} \mb{z}_i \in \mathbbm{R}^{d_h}$. For the graph clustering tasks, the main objective is to group similar nodes into the same cluster, whereas the different nodes will be partitioned into different clusters instead. Therefore, the objective function of graph clustering can be defined as follows:
\begin{equation}
\min_{\bs{\mu}_1, \cdots, \bs{\mu}_l} \min_{\mc{C}} \sum_{j=1}^l \sum_{v_i \in \mc{C}_j} \left \| \mb{z}_i - \bs{\mu}_j \right\|_2.
\end{equation}
The above objective function involves multiple variables to be learned concurrently, which can be trained with the EM algorithm much more effectively instead of error backpropagation. Therefore, instead of re-training the above graph clustering model together with {\our}, we will only take the learned node representations as the node feature input for learning the graph clustering model instead.


\section{Experiments}\label{sec:experiment}

To test the effectiveness of {\our} in learning the graph representations, in this section, we will provide extensive experimental results of {\our} on three real-world benchmark graph datasets, i.e., Cora, Citeseer and Pubmed \cite{YCS16}, respectively.

\noindent \textbf{Reproducibility}. Both the datasets and source code used can be accessed via link\footnote{https://github.com/jwzhanggy/Graph-Bert}. Detailed information about the server used to run the model can be found at the footnote\footnote{GPU Server: ASUS X99-E WS motherboard, Intel Core i7 CPU 6850K@3.6GHz (6 cores), 3 Nvidia GeForce GTX 1080 Ti GPU (11 GB buffer each), 128 GB DDR4 memory and 128 GB SSD swap.}.

\begin{figure}
    \centering
    \begin{subfigure}[b]{.23\textwidth}
    	\includegraphics[width=\linewidth]{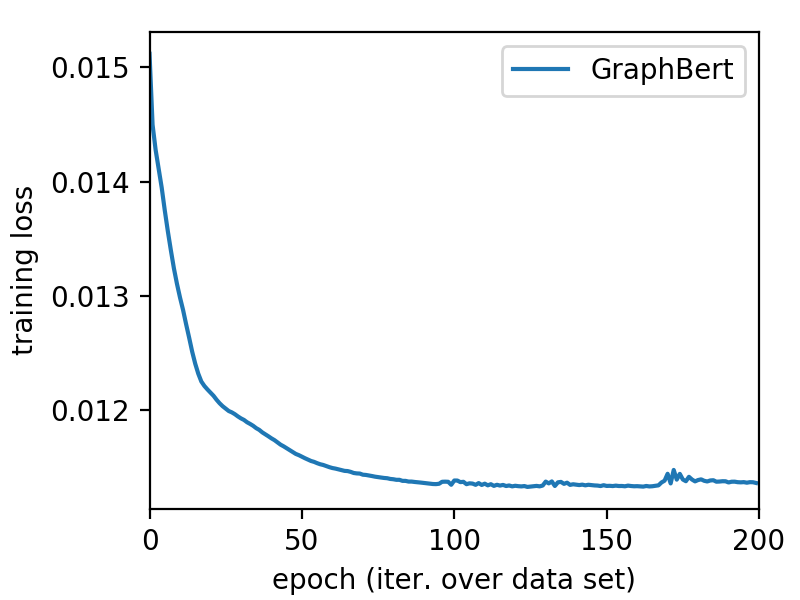}
    	\caption{Node Reconstruction}\label{fig:acc_train}
    \end{subfigure}%
    \hfill
    \begin{subfigure}[b]{.23\textwidth}
    	\includegraphics[width=\linewidth]{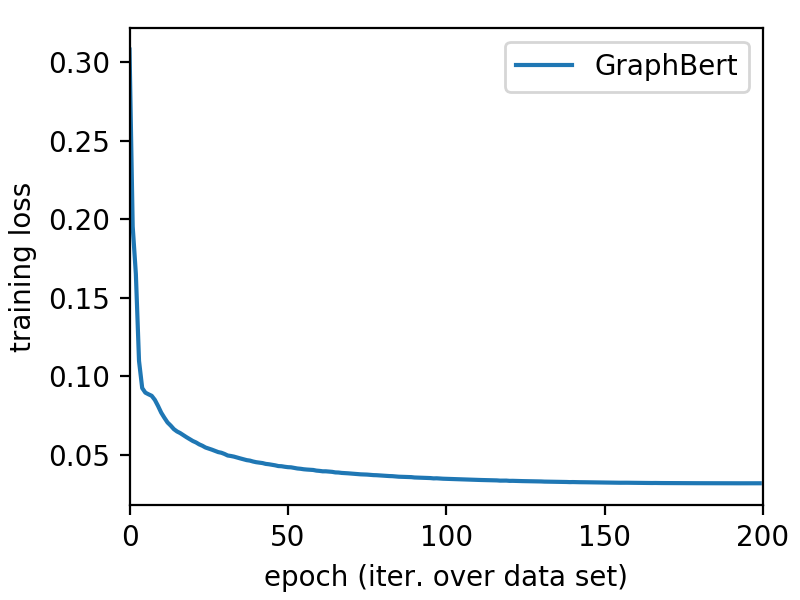}
    	\caption{Graph Recovery}\label{fig:acc_test}
    \end{subfigure}%
    \caption{Pre-training of {\our} on node reconstruction and graph recovery tasks (x axis: iteration; y axis: training loss).}\label{fig:graph_bert_pre_train}
\end{figure}

\begin{figure}[t]
    \centering
    \begin{subfigure}[b]{.23\textwidth}
    	\includegraphics[width=\linewidth]{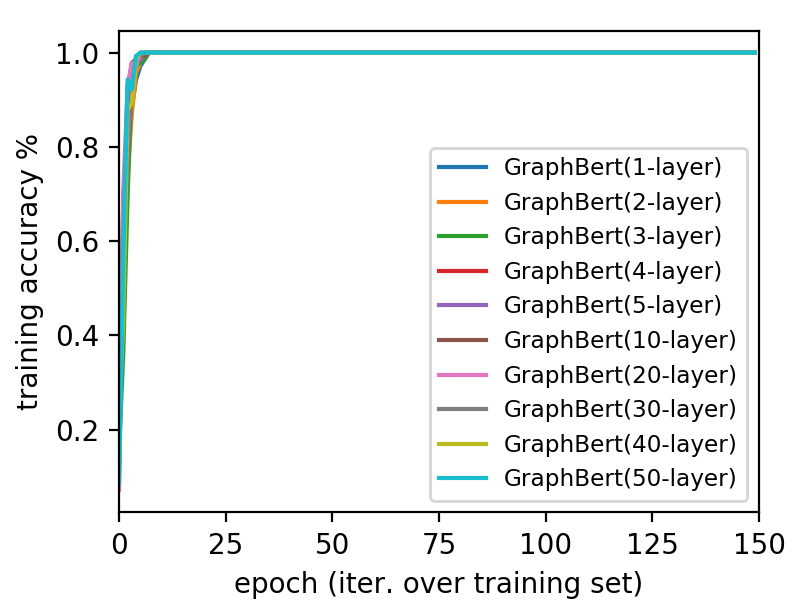}
    	\caption{Training Accuracy}\label{fig:acc_train}
    \end{subfigure}%
    \hfill
    \begin{subfigure}[b]{.23\textwidth}
    	\includegraphics[width=\linewidth]{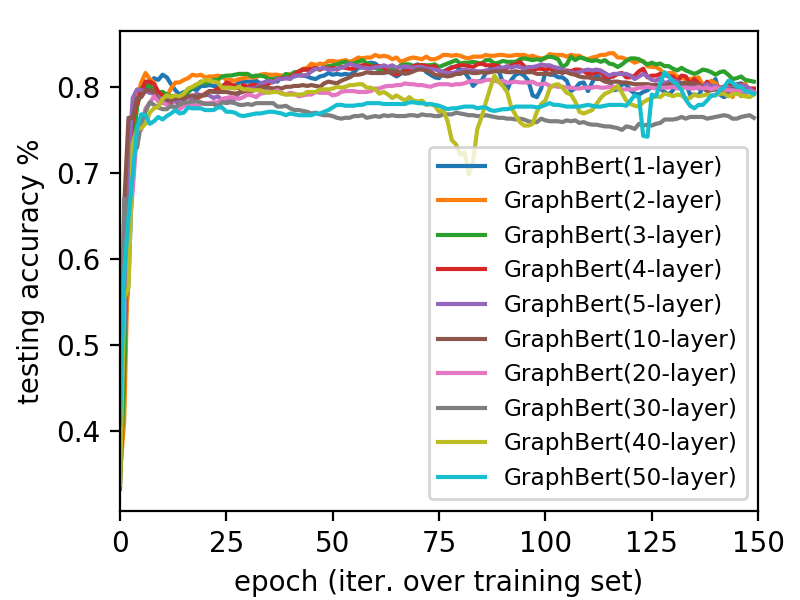}
    	\caption{Testing Accuracy}\label{fig:acc_test}
    \end{subfigure}%
    \caption{The learning performance of {\our} on node classification with 1-layer, $\dotsc$, 5-layer, and 10-layer, $\cdots$, 50-layer on the Cora dataset. The x axis denotes the iterations over the whole training set. The y axes denote the training and testing accuracy, respectively.}\label{fig:graph_bert_acc_analysis}
\end{figure}

\begin{table}[t]
\caption{Learning Performance of {\our} (based on different graph residual terms) compared against existing baseline methods on node classification. The results of {\our} reported here denotes the best observed scores obtained with subgraph size $k \in \{1, 2, \cdots, 10, 15, 20, \cdots, 50\}$.}\label{tab:node_classification}
\centering
\small
\setlength{\tabcolsep}{3pt}
\begin{tabular}{l c c c c }
\toprule
 \multirow{2}{*}{Methods}  & \multicolumn{3}{c}{Datasets (Accuracy)} \\
\cline{2-4}
\addlinespace[0.05cm]
& \textbf{Cora} & \textbf{Citeseer} & \textbf{Pubmed} \\
\hline
\addlinespace[0.05cm]

{LP (\cite{ZGL03}) } &0.680 &0.453 &0.630  \\
{ICA (\cite{LG03})} &0.751  &0.691  &0.739   \\
{ManiReg (\cite{BNS06})} &0.595  &0.601  &0.707   \\
{SemiEmb (\cite{WRC08})} &0.590  &0.596  &0.711  \\
\hline
\addlinespace[0.05cm]
{DeepWalk (\cite{PAS14})} &0.672  &0.432  &0.653   \\
{Planetoid (\cite{YCS16})} &0.757  &0.647  &0.772  \\
{MoNet (\cite{MBMRSB16})} &0.817  &-  &0.788  \\
\hline
\addlinespace[0.05cm]
{{\gcn} (\cite{Kipf_Semi_CORR_16})} &0.815  &0.703  &\textbf{0.790}   \\
{{\gat} (\cite{Velickovic_Graph_ICLR_18})} &\textbf{0.830}  &\textbf{0.725}  &\textbf{0.790}  \\
{{\loopy} (\cite{loopynet})} &\textbf{0.826}  &\textbf{0.716}  &\textbf{0.792}  \\
\hline
\addlinespace[0.05cm]
{\our} &\textbf{0.843}  &\textbf{0.712}  &\textbf{0.793}  \\
\bottomrule
\end{tabular}
\end{table}

\begin{table}[t]
\caption{Analysis of subgraph size $k$ on Cora for model performance (testing accuracy and testing loss) and total time cost.}\label{tab:k_analysis}
 \small
\centering
\setlength{\tabcolsep}{6pt}
\begin{tabular}{|l|c|c|c|}
\hline
 & \multicolumn{3}{|c|}{\textbf{Cora Dataset}} \\
\cline{2-4}
{k }& {\textbf{Test Accuracy}} & {\textbf{Test Loss}} & {\textbf{Total Time Cost} (s)} \\
\hline 
\hline 
1&0.804  &0.791  &\textbf{3.64}   \\
\cline{2-4}
2&0.806  &0.708  &4.02   \\
\cline{2-4}
3&0.819  &0.663  &4.65   \\
\cline{2-4}
4&0.818&0.690  &4.75   \\
\cline{2-4}
5&0.824  &0.636  &5.20   \\
\cline{2-4}
6&0.834 &0.625  &5.62   \\
\cline{2-4}
7&\textbf{0.843} &\textbf{0.620}  &5.96   \\
\cline{2-4}
8&0.828  &0.653  &6.54   \\
\cline{2-4}
9&0.814  &0.679  &6.87   \\
\cline{2-4}
10&0.819 &0.653  &7.26   \\
\cline{2-4}
20&0.819 &0.666  &12.31   \\
\cline{2-4}
30&0.801 &0.710  &17.56   \\
\cline{2-4}
40&0.768 &0.805  &23.77   \\
\cline{2-4}
50&0.759 &0.833  &31.59   \\
\hline
\end{tabular}
\end{table}

\subsection{Dataset and Learning Settings}

The graph benchmark datasets used in the experiments include Cora, Citeseer and Pubmed \cite{YCS16}, which are used in most of the recent state-of-the-art graph neural network research works \cite{Kipf_Semi_CORR_16,Velickovic_Graph_ICLR_18,Zhang2019GResNetGR}. Based on the input graph data, we will first pre-compute the node intimacy scores, based on which subgraph batches will be sampled subject to the subgraph size $k \in \{1, 2, \cdots, 10, 15, 20, \cdots, 50\}$. In addition, we will also pre-compute the node pairwise hop distance and WL node codes. By minimizing the node raw feature reconstruction loss and graph structure recovery loss, {\our} can be effectively pre-trained, whose learned variables will be transferred to the follow-up node classification and graph clustering tasks with/without fine-tuning. In the experiments, we first pre-train {\our} based on the node attribute reconstruction task with 200 epochs, then load and pre-train the same {\our} model again based on the graph structure recovery task with another 200 epochs. In Figure~\ref{fig:graph_bert_pre_train}, we show the learning performance of {\our} on node attribute reconstruction and graph recovery, which converges very fast on both of these tasks.

\begin{table}[t]
\caption{Learning performance of {\our} with different graph residual terms.}\label{tab:graph_residual}
 \small
\centering
\setlength{\tabcolsep}{3.5pt}
\begin{tabular}{|c|l|p{1.2cm}|p{1.2cm}|p{1.2cm}|}
\hline
\multicolumn{2}{|c}{Methods } & \multicolumn{3}{|c|}{Datasets (Accuracy)} \\
\hline
Models & Residual & {\textbf{Cora}} & {\textbf{Citeseer}} & {\textbf{Pubmed}} \\
\hline 
\hline 
\multirow{3}{*}{{\our}}
&none &0.804   &{0.616}  &0.786   \\
\cline{2-5}
&raw&\textbf{0.817}  &\textbf{0.653}  &\textbf{0.786}   \\
\cline{2-5}
&graph-raw&\textbf{0.843}   &\textbf{0.712}  &\textbf{0.793}  \\
\hline
\end{tabular}
\end{table}

\begin{table}[t]
\caption{Learning performance of {\our} with different initial embedding inputs.}\label{tab:embedding_analysis}
 \small
\centering
\setlength{\tabcolsep}{3.5pt}
\begin{tabular}{|c|l|p{1.2cm}|p{1.2cm}|p{1.2cm}|}
\hline
\multicolumn{2}{|c}{Methods } & \multicolumn{3}{|c|}{Datasets (Accuracy)} \\
\hline
Models & Embedding & {\textbf{Cora}} & {\textbf{Citeseer}} & {\textbf{Pubmed}} \\
\hline 
\hline 
\multirow{5}{*}{{\our}}
&hop distance&{0.307}   &0.348  &{0.445}  \\
\cline{2-5}
&position&{0.323}   &0.331  &{0.395}  \\
\cline{2-5}
&wl role&0.457   &{0.345}  &0.443   \\
\cline{2-5}
&raw feature&\textbf{0.795}  &\textbf{0.611}  &\textbf{0.780}   \\
\cline{2-5}
&all&\textbf{0.804}   &\textbf{0.616}  &\textbf{0.786}  \\
\hline
\end{tabular}
\end{table}

\begin{table}[t]
\caption{Clustering results of {\our} without pre-training solely based on node raw features (MI: mutual information).}\label{tab:graph_clustering}
\centering
\small
\setlength{\tabcolsep}{3pt}
\begin{tabular}{l c c c c }
\toprule
 \multirow{2}{*}{Metrics}  & \multicolumn{3}{c}{Datasets} \\
\cline{2-4}
\addlinespace[0.05cm]
& \textbf{Cora} & \textbf{Citeseer} & \textbf{Pubmed} \\
\hline
\addlinespace[0.05cm]

{Rand} &0.080  & 0.249  &0.281   \\
{Adjusted MI} &0.130 &0.287 &0.313  \\
{Normalized MI} &0.133  &0.289  &0.313   \\
{Homogeneity} &0.133  &0.287  &0.280  \\
{Completeness} &0.132  &0.291  &0.355  \\
\bottomrule
\end{tabular}
\end{table}

\subsubsection{Default Parameter Settings}

If not clearly specified, the results reported in this paper are based on the following parameter settings of {\our}: \textit{subgraph size}: $k=7$ (Cora), $k=5$ (Citeseer) and $k=30$ (Pubmed); \textit{hidden size}: 32; \textit{attention head number}: 2; \textit{hidden layer number}: $D=2$; \textit{learning rate}: 0.01 (Cora) and 0.001 (Citeseer) and 0.0005 (Pubmed); \textit{weight decay}: $5e^{-4}$; \textit{intermediate size}: 32; \textit{hidden dropout rate}: 0.5; \textit{attention dropout rate}: 0.3; \textit{graph residual term}: graph-raw; \textit{training epoch}: 150 (Cora), 500 (Pubmed), 2000 (Citeseer).

\subsection{Node Classification without Pre-training}

{\our} is a powerful mode and it can be applied to address various graph learning tasks in the standalone mode. To show the effectiveness of {\our}, we will first provide the experimental results of {\our} on the node classification task without pre-training here, whereas the pre-trained {\our} based node classification results will be provided in Section~\ref{subsec:pre_train_result} in more detail. Here, we will follow the identical train/validation/test set partitions used in the existing graph neural network papers \cite{YCS16} for fair comparisons.

\subsubsection{Learning Convergence of Deep {\our}}

In Figure~\ref{fig:graph_bert_acc_analysis}, we illustrate the training records of {\our} for node classification on the Cora dataset. To show that {\our} is different from other GNN models and {\our} works with deep architectures, we also change the model depth with values from $\{1, 2, \cdots, 5, 10, 20, \cdot, 50\}$. According to the plots, {\our} can converge very fast (with less than 10 epochs) on the training set. What's more, as the model depth increases, {\our} will not suffer from the suspended animation problem. Even the very deep {\our} (50 layers) can still respond effectively to the training data and achieve good learning performance.

\subsubsection{Main Results}

The learning results of {\our} (with different graph residual terms) on node classification are provided in Table~\ref{tab:node_classification}. The comparison methods used here cover both classic and state-of-the-art GNN models. For the variant models which extend {\gcn} and {\gat} (with new learning settings, include more training data, re-configure the graph structure or use new optimization methods), we didn't compare them here. However, similar techniques proposed by these extension works can be used to further help improve {\our} as well. According to the achieved scores, we observe that {\our} can out-perform most of these baseline methods with a big improvement on both Cora and Pubmed. On Citeseer, its perofrmance is also among the top 3.

\begin{table*}[t]
\caption{Performance comparison of {\our} on fine-tuning tasks with/without pre-training. For all the models shown here, we will only use $\frac{1}{5}$ of the normal training max epochs as used by {\our} in Table~\ref{tab:node_classification}. For KMeans, the epoch denotes its max-iter parameter.}\label{tab:performance_summary}
\centering
\setlength{\tabcolsep}{3.5pt}
\begin{tabular}{|c|l|p{0.9cm}|p{1.2cm}|p{0.9cm}|p{1.2cm}|p{0.9cm}|p{1.2cm}| }
\hline
\multicolumn{2}{|c}{Methods } & \multicolumn{6}{|c|}{Datasets (Accuracy/Rand \& Epoch)} \\
\hline
Pre-Train Task & Fine-Tune Task & \multicolumn{2}{|c}{\textbf{Cora}} & \multicolumn{2}{|c}{\textbf{Citeseer}} & \multicolumn{2}{|c|}{\textbf{Pubmed}} \\
\hline 
\hline 
\multirow{2}{*}{Node Reconstruction}
&Node Classification&0.827 &30  &0.649 &400 &0.780 &100 \\
\cline{2-8}
&Graph Clustering&\textbf{0.400} &30  &\textbf{0.312} &400 &0.027 &100 \\
\hline
\hline
\multirow{2}{*}{Structural Recovery}
&Node Classification&0.823 &30  &0.662 &400 &0.788 &100 \\
\cline{2-8}
&Graph Clustering&0.123 &30  &0.090 &400 &0.132 &100 \\
\hline
\hline
\multirow{2}{*}{Both}
&Node Classification&\textbf{0.836} &30  &\textbf{0.672} &400 &\textbf{0.791} &100 \\
\cline{2-8}
&Graph Clustering&0.177 &30  &0.203 &400 &0.159 & 100 \\
\hline
\hline
\multirow{2}{*}{None}
&Node Classification&0.805 &30  &0.654 &400&0.786 &100 \\
\cline{2-8}
&Graph Clustering&0.080 &30  &0.249 &400 &\textbf{0.281} &100 \\
\hline
\end{tabular}
\end{table*}

\subsubsection{Subgraph Size $k$ Analysis}

As illustrated in Table~\ref{tab:k_analysis}, we provide the learning performance analysis of {\our} with different subgraph sizes, i.e., parameter $k$, on the Cora dataset. According to the results, parameter $k$ affects the learning performance of {\our} a lot, since it defines how many nearby nodes will be used to define the nodes' learning context. For the Cora dataset, we observe that the learning performance of {\our} improves steadily as $k$ increases from $1$ to $7$. After that, as $k$ further increases, the performance will degrade dramatically. For the good scores with $k=1$, partial contributions come from the graph residual terms in {\our}. The time cost of {\our} increases as $k$ goes larger, which is very minor actually compared with other existing GNN models, like {\gcn} and {\gat}. Similar results can be observed for the other two datasets, but the optimal $k$ are different.

\subsubsection{Graph Residual Analysis}

What's more, in Table~\ref{tab:graph_residual}, we also provide the learning results of {\our} with different graph residual terms. According to the scores, {\our} with graph-raw residual term can outperform the other two, which is also consistent with the experimental observations on these different residual terms as reported in \cite{Zhang2019GResNetGR}.

\subsubsection{Initial Embedding Analysis}

As shown in Table~\ref{tab:embedding_analysis}, we provide the learning performance of {\our} on these three datasets, which takes different initial embeddings as the input. To better show the performance differences, the {\our} used here doesn't involve any residual learning. According to the results, using the \textit{Weisfeiler-Lehman role embedding}, \textit{hop based distance embedding} and \textit{intimacy based positional embedding} vectors along, {\our} cannot work very well actually, whereas the raw feature embeddings do contribute a lot. Meanwhile, by incorporating such complementary embeddings into the raw feature embedding, the model can achieve better performance than using raw feature embedding only.

\subsection{Graph Clustering without Pre-Training}

In Table~\ref{tab:graph_clustering}, we show the learning results of {\our} on graph clustering without any pre-training on the three datasets. Formally, the clustering component used in {\our} is KMeans, which takes the nodes' raw feature vectors as the input. The results are evaluated with several different metrics shown above.

\subsection{Pre-training vs. No Pre-training}\label{subsec:pre_train_result}

The results reported in the previous subsections are all based on the {\our} without pre-training actually. Here, we will provide the experimental results on {\our} with pre-training to show their differences. According to the experiments, given enough training epochs, models with/without pre-training can both converge to very good learning results. Therefore, to highlight the differences, we will only use $\frac{1}{5}$ of the normal training epochs here for fine-tuning {\our}, and the results are provided in Table~\ref{tab:performance_summary}. We also show the performance of {\our} without pre-training here for comparison.

According to the scores, for most of the datasets, pre-training do give {\our} a good initial state, which helps the model achieve better performance with only a very small number of fine-tuning epochs. On Cora and Citeseer, pre-training helps both the node classification and graph clustering tasks. Meanwhile, for Pubmed, pre-training helps node classification but degrades the results on graph clustering. Also pre-training with both node classification and graph recovery help the model to capture more information from the graph data, which also lead to higher scores than the models with single pre-training tasks.

\section{Conclusion}\label{sec:conclusion}

In this paper, we have introduced the new {\our} model for graph representation learning. Different from existing GNNs, {\our} works well in deep architectures and will not suffer from the common problems with other GNNs. Based on a batch of linkless subgraphs sampled from the original graph data, {\our} can effectively learn the representations of the target node with the extended graph-transformer layers introduced in this paper. {\our} can serve as the graph representation learning component in graph learning pipeline. The pre-trained {\our} can be transferred and applied to address new tasks either directly or with necessary fine-tuning.

{
\bibliographystyle{named}
\bibliography{reference}

\begin{thebibliography}{}

\bibitem[\protect\citeauthoryear{Adamic and Adar}{2003}]{adamic2003friends}
Eytan Adamic and Lada~A. Adar.
\newblock Friends and neighbors on the web.
\newblock (3):211--230, July 2003.

\bibitem[\protect\citeauthoryear{Belkin \bgroup \em et al.\egroup
  }{2006}]{BNS06}
Mikhail Belkin, Partha Niyogi, and Vikas Sindhwani.
\newblock Manifold regularization: A geometric framework for learning from
  labeled and unlabeled examples.
\newblock {\em J. Mach. Learn. Res.}, 7:2399?2434, December 2006.

\bibitem[\protect\citeauthoryear{Chung \bgroup \em et al.\egroup
  }{2014}]{DBLP:journals/corr/ChungGCB14}
Junyoung Chung, {\c{C}}aglar G{\"{u}}l{\c{c}}ehre, KyungHyun Cho, and Yoshua
  Bengio.
\newblock Empirical evaluation of gated recurrent neural networks on sequence
  modeling.
\newblock {\em CoRR}, abs/1412.3555, 2014.

\bibitem[\protect\citeauthoryear{Dai \bgroup \em et al.\egroup
  }{2019}]{DBLP:journals/corr/abs-1901-02860}
Zihang Dai, Zhilin Yang, Yiming Yang, Jaime~G. Carbonell, Quoc~V. Le, and
  Ruslan Salakhutdinov.
\newblock Transformer-xl: Attentive language models beyond a fixed-length
  context.
\newblock {\em CoRR}, abs/1901.02860, 2019.

\bibitem[\protect\citeauthoryear{Devlin \bgroup \em et al.\egroup
  }{2018}]{DCLT18}
Jacob Devlin, Ming{-}Wei Chang, Kenton Lee, and Kristina Toutanova.
\newblock {BERT:} pre-training of deep bidirectional transformers for language
  understanding.
\newblock {\em CoRR}, abs/1810.04805, 2018.

\bibitem[\protect\citeauthoryear{Grover and
  Leskovec}{2016}]{DBLP:journals/corr/GroverL16}
Aditya Grover and Jure Leskovec.
\newblock node2vec: Scalable feature learning for networks.
\newblock {\em CoRR}, abs/1607.00653, 2016.

\bibitem[\protect\citeauthoryear{G{\"u}rel \bgroup \em et al.\egroup
  }{2019}]{Merve_An_19}
Nezihe~Merve G{\"u}rel, Hansheng Ren, Yujing Wang, Hui Xue, Yaming Yang, and
  Ce~Zhang.
\newblock An anatomy of graph neural networks going deep via the lens of mutual
  information: Exponential decay vs. full preservation.
\newblock {\em ArXiv}, abs/1910.04499, 2019.

\bibitem[\protect\citeauthoryear{Hamilton \bgroup \em et al.\egroup
  }{2017}]{Hamilton_Inductive_17}
William~L. Hamilton, Rex Ying, and Jure Leskovec.
\newblock Inductive representation learning on large graphs.
\newblock {\em CoRR}, abs/1706.02216, 2017.

\bibitem[\protect\citeauthoryear{Hammond \bgroup \em et al.\egroup
  }{2011}]{Hammond_2011}
David~K. Hammond, Pierre Vandergheynst, and Remi Gribonval.
\newblock Wavelets on graphs via spectral graph theory.
\newblock {\em Applied and Computational Harmonic Analysis}, 30(2):129?150, Mar
  2011.

\bibitem[\protect\citeauthoryear{He \bgroup \em et al.\egroup
  }{2018}]{DBLP:journals/corr/abs-1811-08883}
Kaiming He, Ross~B. Girshick, and Piotr Doll{\'{a}}r.
\newblock Rethinking imagenet pre-training.
\newblock {\em CoRR}, abs/1811.08883, 2018.

\bibitem[\protect\citeauthoryear{Hochreiter and
  Schmidhuber}{1997}]{Hochreiter_Long_Neural_97}
Sepp Hochreiter and J\"{u}rgen Schmidhuber.
\newblock Long short-term memory.
\newblock {\em Neural Comput.}, 9(8), November 1997.

\bibitem[\protect\citeauthoryear{Huang and Carley}{2019}]{Huang_Inductive_19}
Binxuan Huang and Kathleen~M. Carley.
\newblock Inductive graph representation learning with recurrent graph neural
  networks.
\newblock {\em CoRR}, abs/1904.08035, 2019.

\bibitem[\protect\citeauthoryear{Jaccard}{1901}]{jaccard1901etude}
Paul Jaccard.
\newblock \'{E}tude comparative de la distribution florale dans une portion des
  alpes et des jura.
\newblock {\em Bulletin del la Soci\'{e}t\'{e} Vaudoise des Sciences
  Naturelles}, 37:547--579, 1901.

\bibitem[\protect\citeauthoryear{Jin \bgroup \em et al.\egroup
  }{2018}]{Jin_Junction_18}
Wengong Jin, Regina Barzilay, and Tommi~S. Jaakkola.
\newblock Junction tree variational autoencoder for molecular graph generation.
\newblock {\em CoRR}, abs/1802.04364, 2018.

\bibitem[\protect\citeauthoryear{Katz}{1953}]{Katz1953}
Leo Katz.
\newblock A new status index derived from sociometric analysis.
\newblock {\em Psychometrika}, 18(1):39--43, Mar 1953.

\bibitem[\protect\citeauthoryear{Kim}{2014}]{kim-2014-convolutional}
Yoon Kim.
\newblock Convolutional neural networks for sentence classification.
\newblock In {\em Proceedings of the 2014 Conference on Empirical Methods in
  Natural Language Processing ({EMNLP})}, pages 1746--1751, Doha, Qatar,
  October 2014. Association for Computational Linguistics.

\bibitem[\protect\citeauthoryear{Kipf and Welling}{2016}]{Kipf_Semi_CORR_16}
Thomas~N. Kipf and Max Welling.
\newblock Semi-supervised classification with graph convolutional networks.
\newblock {\em CoRR}, abs/1609.02907, 2016.

\bibitem[\protect\citeauthoryear{Klicpera \bgroup \em et al.\egroup
  }{2018}]{DBLP:journals/corr/abs-1810-05997}
Johannes Klicpera, Aleksandar Bojchevski, and Stephan G{\"{u}}nnemann.
\newblock Personalized embedding propagation: Combining neural networks on
  graphs with personalized pagerank.
\newblock {\em CoRR}, abs/1810.05997, 2018.

\bibitem[\protect\citeauthoryear{Lan \bgroup \em et al.\egroup
  }{2019}]{lan2019albert}
Zhenzhong Lan, Mingda Chen, Sebastian Goodman, Kevin Gimpel, Piyush Sharma, and
  Radu Soricut.
\newblock Albert: A lite bert for self-supervised learning of language
  representations, 2019.

\bibitem[\protect\citeauthoryear{Li \bgroup \em et al.\egroup
  }{2018}]{Li_Deeper_CORR_18}
Qimai Li, Zhichao Han, and Xiao{-}Ming Wu.
\newblock Deeper insights into graph convolutional networks for semi-supervised
  learning.
\newblock {\em CoRR}, abs/1801.07606, 2018.

\bibitem[\protect\citeauthoryear{Lu and Getoor}{2003}]{LG03}
Qing Lu and Lise Getoor.
\newblock Link-based classification.
\newblock In {\em Proceedings of the Twentieth International Conference on
  International Conference on Machine Learning}, ICML'03, page 496?503. AAAI
  Press, 2003.

\bibitem[\protect\citeauthoryear{Meng and Zhang}{2019}]{Meng_Isomorphic_19}
Lin Meng and Jiawei Zhang.
\newblock Isonn: Isomorphic neural network for graph representation learning
  and classification.
\newblock {\em CoRR}, abs/1907.09495, 2019.

\bibitem[\protect\citeauthoryear{Monti \bgroup \em et al.\egroup
  }{2016}]{MBMRSB16}
Federico Monti, Davide Boscaini, Jonathan Masci, Emanuele Rodol{\`{a}}, Jan
  Svoboda, and Michael~M. Bronstein.
\newblock Geometric deep learning on graphs and manifolds using mixture model
  cnns.
\newblock {\em CoRR}, abs/1611.08402, 2016.

\bibitem[\protect\citeauthoryear{Niepert \bgroup \em et al.\egroup
  }{2016}]{DBLP:journals/corr/NiepertAK16}
Mathias Niepert, Mohamed Ahmed, and Konstantin Kutzkov.
\newblock Learning convolutional neural networks for graphs.
\newblock {\em CoRR}, abs/1605.05273, 2016.

\bibitem[\protect\citeauthoryear{Perozzi \bgroup \em et al.\egroup
  }{2014a}]{Perozzi:2014:DOL:2623330.2623732}
Bryan Perozzi, Rami Al-Rfou, and Steven Skiena.
\newblock Deepwalk: Online learning of social representations.
\newblock In {\em Proceedings of the 20th ACM SIGKDD International Conference
  on Knowledge Discovery and Data Mining}, KDD '14, pages 701--710, New York,
  NY, USA, 2014. ACM.

\bibitem[\protect\citeauthoryear{Perozzi \bgroup \em et al.\egroup
  }{2014b}]{PAS14}
Bryan Perozzi, Rami Al{-}Rfou, and Steven Skiena.
\newblock Deepwalk: Online learning of social representations.
\newblock {\em CoRR}, abs/1403.6652, 2014.

\bibitem[\protect\citeauthoryear{Shang \bgroup \em et al.\egroup
  }{2019}]{DBLP:journals/corr/abs-1906-00346}
Junyuan Shang, Tengfei Ma, Cao Xiao, and Jimeng Sun.
\newblock Pre-training of graph augmented transformers for medication
  recommendation.
\newblock {\em CoRR}, abs/1906.00346, 2019.

\bibitem[\protect\citeauthoryear{Sun \bgroup \em et al.\egroup
  }{2019}]{sun2019adagcn}
Ke~Sun, Zhouchen Lin, and Zhanxing Zhu.
\newblock Adagcn: Adaboosting graph convolutional networks into deep models,
  2019.

\bibitem[\protect\citeauthoryear{Ugander \bgroup \em et al.\egroup
  }{2011}]{Ugander_Anatomy_11}
Johan Ugander, Brian Karrer, Lars Backstrom, and Cameron Marlow.
\newblock The anatomy of the facebook social graph.
\newblock {\em CoRR}, abs/1111.4503, 2011.

\bibitem[\protect\citeauthoryear{Vaswani \bgroup \em et al.\egroup
  }{2017}]{VSPUJGKP17}
Ashish Vaswani, Noam Shazeer, Niki Parmar, Jakob Uszkoreit, Llion Jones,
  Aidan~N. Gomez, Lukasz Kaiser, and Illia Polosukhin.
\newblock Attention is all you need.
\newblock {\em CoRR}, abs/1706.03762, 2017.

\bibitem[\protect\citeauthoryear{Veli{\v{c}}kovi{\'{c}} \bgroup \em et
  al.\egroup }{2018}]{Velickovic_Graph_ICLR_18}
Petar Veli{\v{c}}kovi{\'{c}}, Guillem Cucurull, Arantxa Casanova, Adriana
  Romero, Pietro Li{\`{o}}, and Yoshua Bengio.
\newblock {Graph Attention Networks}.
\newblock {\em International Conference on Learning Representations}, 2018.

\bibitem[\protect\citeauthoryear{Weston \bgroup \em et al.\egroup
  }{2008}]{WRC08}
Jason Weston, Fr\'{e}d\'{e}ric Ratle, and Ronan Collobert.
\newblock Deep learning via semi-supervised embedding.
\newblock In {\em Proceedings of the 25th International Conference on Machine
  Learning}, ICML'08, page 1168?1175, New York, NY, USA, 2008. Association for
  Computing Machinery.

\bibitem[\protect\citeauthoryear{Yang \bgroup \em et al.\egroup }{2016}]{YCS16}
Zhilin Yang, William~W. Cohen, and Ruslan Salakhutdinov.
\newblock Revisiting semi-supervised learning with graph embeddings.
\newblock {\em CoRR}, abs/1603.08861, 2016.

\bibitem[\protect\citeauthoryear{Zhang and Meng}{2019}]{Zhang2019GResNetGR}
Jiawei Zhang and Lin Meng.
\newblock Gresnet: Graph residual network for reviving deep gnns from suspended
  animation.
\newblock {\em ArXiv}, abs/1909.05729, 2019.

\bibitem[\protect\citeauthoryear{Zhang \bgroup \em et al.\egroup
  }{2018}]{ZCG18}
Jiawei Zhang, Limeng Cui, and Fisher~B. Gouza.
\newblock {SEGEN:} sample-ensemble genetic evolutional network model.
\newblock {\em CoRR}, abs/1803.08631, 2018.

\bibitem[\protect\citeauthoryear{Zhang}{2018}]{loopynet}
Jiawei Zhang.
\newblock Deep loopy neural network model for graph structured data
  representation learning.
\newblock {\em CoRR}, abs/1805.07504, 2018.

\bibitem[\protect\citeauthoryear{Zhu \bgroup \em et al.\egroup }{2003}]{ZGL03}
Xiaojin Zhu, Zoubin Ghahramani, and John Lafferty.
\newblock Semi-supervised learning using gaussian fields and harmonic
  functions.
\newblock In {\em Proceedings of the Twentieth International Conference on
  International Conference on Machine Learning}, ICML'03, page 912?919. AAAI
  Press, 2003.

\end{thebibliography}
}


\end{document}